\documentclass[letterpaper, 10 pt, conference]{ieeeconf}  

\usepackage{graphicx}
\usepackage{graphicx}
\usepackage[export]{adjustbox}
\usepackage{setspace}
\usepackage{subfigure}
\usepackage{mathptmx}
\usepackage{bm}
\usepackage{amsmath,amssymb} 
\usepackage{color}
\usepackage{algorithm}
\usepackage{algpseudocode}
\algnewcommand{\LineComment}[1]{\State \(\triangleright\) #1}
\usepackage{enumerate}
\usepackage{epstopdf}
\usepackage{array}  
\usepackage{eqparbox}
\usepackage{multirow}
\usepackage{url}
\usepackage{mathtools}
\usepackage{hyperref}
\usepackage{adjustbox}
\usepackage{caption}
\usepackage{textcomp}

\usepackage[T1]{fontenc}
\usepackage[font=small,labelfont=bf,tableposition=top]{caption}

\usepackage{array}
\usepackage{mathtools,eqparbox}
\usepackage{tikz}
\usepackage{lipsum}
\usepackage{float}

\usepackage{amsmath} 
\usepackage{subfigure}
\usepackage{epstopdf}
\usepackage{url}
\usepackage{multirow}
\usepackage{algorithm}
\usepackage{algorithmicx}
\usepackage{media9}
\usepackage{multimedia}
\usepackage{subfigure}
\usepackage{algpseudocode}
\usepackage{amsmath} 
\usepackage{amssymb}
\usepackage{graphicx}

\IEEEoverridecommandlockouts                              

\title{\LARGE \bf
Control Barrier Function-based Predictive Control for Close Proximity operation of UAVs inside a Tunnel
}

\author{Vedant Mundheda$^{1}$, Damodar Datta K$^{1}$ and Harikumar Kandath$^{1}$
\thanks{$^{1}$Robotics Research Center, IIIT-Hyderabad vedant.mundheda@research.iiit.ac.in, damodardatta@gmail.com, harikumar.k@iiit.ac.in}%
}

\begin{document}

\maketitle
\thispagestyle{empty}
\pagestyle{empty}

\begin{abstract}

This paper introduces a method for effectively controlling the movement of an Unmanned Aerial Vehicle (UAV) within a tunnel. The primary challenge of this problem lies in the UAV's exposure to nonlinear distance-dependent torques and forces generated by the tunnel walls, along with the need to operate safely within a defined region while in close proximity to these walls. To address this problem, the paper proposes the implementation of a Model Predictive Control (MPC) framework with constraints based on Control Barrier Function (CBF). The paper approaches the issue in two distinct ways; first, by maintaining a safe distance from the tunnel walls to avoid the effects of both the walls and ceiling, and second, by minimizing the distance from the walls to effectively manage the nonlinear forces associated with close proximity tasks. Finally, the paper demonstrates the effectiveness of its approach through testing on simulation for various close proximity trajectories with the realistic model of aerodynamic disturbances due to the proximity of the ceiling and boundary walls.
\end{abstract}

\section{INTRODUCTION} 

Present times have witnessed the widespread deployment of Unmanned Aerial Vehicles (UAVs) in a variety of domains, ranging from delivery, search, and rescue to monitoring \cite{droneapplication}. Certain civil inspection and delivery tasks necessitate close-range operations near stationary obstructions, such as bridges and buildings \cite{civilexample}. Furthermore, UAV based indoor missions involving inspection of tunnels, rooms, aircraft fuel tanks, coal mines and AC ducts, offer significant advantages over traditional manual methods by reducing the time and effort required while also minimizing risks to human safety. Nonetheless, when conducting inspection tasks in close proximity to obstacles or walls, the UAV's aerial dynamics are subject to various force and torque disturbances, leading to potential instability and safety concerns. To account for such disturbances from all directions, we demonstrate our controller for operating inside a tunnel.

The behavior of a UAV as it approaches the walls of a tunnel is characterized by nonlinear variation in its thrust, attributable to the intricate aerodynamic interactions at play \cite{gecesw}. As a result, a region of operation that is deemed unsafe can be identified in the vicinity of the wall or obstacle, necessitating the confinement of the UAV to a remaining safe region. Nonetheless, certain inspection tasks may require the UAV to operate in close proximity to the wall. Consequently, the controller must be designed to facilitate stability in the presence of such nonlinear disturbances. \cite{multirotor_forces} demonstrates the safe distance for operation is beyond $2 \times$  \textit{Radius of Propeller} from the obstruction or wall.

\begin{figure}
    \centering
    \includegraphics[width=0.5\textwidth]{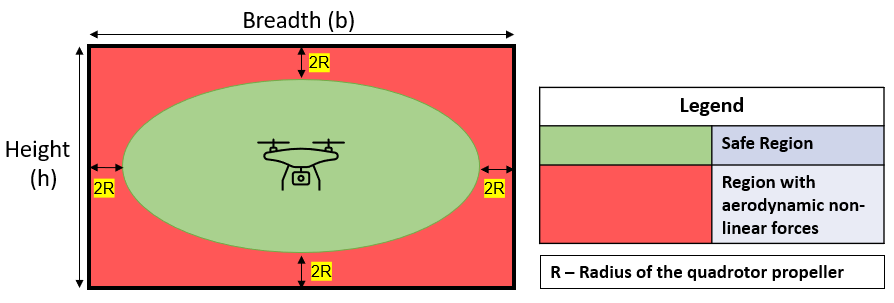}
    \caption{Depicts operation of the UAV in a safe region with minimal aerodynamic effects from the wall. If the UAV goes closer than $2 \times$ R from the walls, it experiences turbulent forces, which tend to destabilize the UAV and cause collision.}
    \label{fig1}
\end{figure}

\section{RELATED WORK}

The literature is sufficiently populated with efforts to model ceiling and ground effects \cite{gecesw,multirotor_forces,ge,ce_model}, but there is a clear gap in formulating control algorithms to tackle these effects in a combined fashion.    
Nonlinear Model Predictive Control (MPC) \cite{9145644} has been used for navigation and obstacle avoidance of UAVs for real-time utilities. MPC provides the predictive ability \cite{8594266} which aids in performing agile maneuvers with high precision and smooth control actions. \cite{9547384} tries to limit the risk of unsafety by formulating a probabilistic guarantee, but fails to provide a rigid safety guarantee to avoid obstacles. \cite{9938397} utilises partial sensor information to navigate through unknown environments by providing partial safety guarantees.  

Control Barrier Function (CBF) \cite{8796030} is used to guarantee safety-critical control for various domains, including dynamic robotic systems. \cite{8796030} introduces safety, safety sets, and describes using CBF to enforce safety in a minimally invasive fashion by not increasing the control effort or trajectory cost. CBF has been used as a constraint to MPC \cite{8887800} to provide safety guarantees while addressing the case of conflict between safety and performance. This provides improved performance to MPC while providing safety guarantees. \cite{9655081} shows collision avoidance for multi UAV swarm to reach desired locations and providing safety guarantees. \cite{mundheda2022predictive} utilizes MPC while handling external wind disturbances. 
Although nonlinear controllers have been tried separately for ground and ceiling effects, no effort has been made to minimize the impacts of these disruptions using a disturbance resistive barrier function.

\section{CONTRIBUTIONS}

The paper contributes in the following ways:

\begin{enumerate}
    \item To the best of the author's knowledge, this paper marks the initial endeavor to address the challenges of ground, ceiling, and wall effects simultaneously in a closed space for a UAV via the utilization of a model predictive controller.
    \item This paper also proposes the use of CBF as a bounding function to bound the UAV into the Safe region (in Fig. \ref{fig1}) to prevent interaction with the aerodynamical forces of tunnel effect.
    \item The contemporary CBF function is modified to tackle disturbances and provide safety guarantees in the presence of bounded external disturbances.
\end{enumerate}

The paper follows the structure with Section IV describing the UAV dynamic, a conventional CBF, and different aerodynamic effects acting on the UAV. Section V provides the problem formulation, and Section VI describes the outer loop MPC with CBF constraints and inner loop PID. Section VII explains the simulation results for different cases, and Section VIII concludes the paper. 
    
\section{PRELIMINARIES}

\subsection{UAV Dynamics}

\par UAV translational dynamics are given in (\ref{eq1}).
\begin{equation}
\!
\begin{aligned}
\ddot{\mathbf{p}} &= \mathbf{g} + {^{I}R_B}\mathbf{T}/m 
\end{aligned}
\label{eq1}
\end{equation}
where $\mathbf{p}$ is the position of the center of the UAV in the inertial frame, $m$ is the mass of the UAV, and $T$ is the thrust vector acting on the UAV in the body frame. $^{I}R_{B}$ denotes standard rotation matrix in 3D for transformation from frame $B$ to frame $I$ \cite{Rotation_matrix}.

The rotational dynamics are given in (\ref{eq2}) where the angular acceleration in the body frame is ${\mathbf{\omega}}$.
\begin{equation}
\dot{\mathbf{\omega}} = \mathbf{I}^{-1}(\mathbf{\tau} - {{\mathbf{\omega}}} \times \mathbf{I}{\mathbf{{\omega}}})
\label{eq2}
\end{equation}

where $\tau$ and $\mathbf{I}$ are, the torque acting on the UAV and inertia matrix defined in the body frame.

\par The combined UAV dynamics using (\ref{eq1}) and (\ref{eq2}) is presented below in matrix form.

\begin{equation}
\label{eq3}
\resizebox{\hsize}{!}{$
\begin{bmatrix}
m\mathbf{I}_{3\times3} && \mathbf{0}_{3\times3} \\
\mathbf{0}_{3\times3} && \mathbf{I} \\
\end{bmatrix} 
\begin{bmatrix}
\ddot{\mathbf{p}} \\
\dot{\mathbf{\omega}} \\
\end{bmatrix} +
\begin{bmatrix}
\mathbf{0}_{3\times3} \\
\mathbf{{\omega}} \times \mathbf{I}\mathbf{{\omega}} \\
\end{bmatrix} 
=
\begin{bmatrix}
m\mathbf{g} + {^{I}R_B}T \\ 
\mathbf{\tau} \\
\end{bmatrix}$}
\end{equation} 

where $\mathbf{I}$ denotes identity and $\mathbf{0}$ denotes a null matrix. 

\subsection{Control Barrier Function}

Dynamics of a UAV in control affine form are given in (\ref{eq_control_affine}):

\begin{equation}
\label{eq_control_affine}
\mathbf{\dot x} = \mathbf{A} \mathbf{x} + \mathbf{B} \mathbf{u}
\end{equation}

where $A$ is a square matrix of dimension $4$x$4$ and $B$ is a matrix of dimension $4$x$2$.

$h(\mathbf{x})$ is a valid CBF if it is differentiable and follows the conditions in (\ref{barrier_inequality}).
\begin{equation}
  \left\{\begin{array}{@{}l@{}}
    h(\mathbf{x}) >  0 , \ \forall \ \mathbf{x}  \in  \zeta\\
    h(\mathbf{x}) =  0 , \  \forall  \ \mathbf{x}  \in  \delta\zeta\\
  \end{array}\right.\,
  \label{barrier_inequality}
\end{equation}

where $\zeta$ is the set of all states of the UAV which lie in the safe region and $\delta\zeta$ are the states of the UAV on the boundary of the safe region. 

If the UAV is initially located within the secure area $\zeta$, the principle of forward invariance can be applied by verifying that $\dot h(\mathbf{x}) \geq 0$. This principle ensures that the UAV remains within the safe region if it commences within it. To enhance optimization for ideal trajectory tracking while also providing safety guarantees, this principle can be extended to an invariance condition where $\dot h(\mathbf{x}) \geq -\gamma h(\mathbf{x})$. This invariance condition induces the asymptotic convergence of $h(\mathbf{x})$ to 0. The condition for invariance is presented in equation (\ref{invariance}).

\begin{equation}
    \frac{\partial h(\mathbf{x})}{\partial x} (\mathbf{A} \mathbf{x} + \mathbf{B} \mathbf{u}) + \gamma h^{z}(\mathbf{x}) \geq 0 
    \label{invariance}
\end{equation}

where $\gamma > 0$ is the relaxation coefficient and $z > 0$ is the exponential limit of convergence for the CBF. We define CBF to avoid point obstacles as $h$ in (\ref{CBF_point}).

\begin{equation}
    h(\mathbf{x}) = \sqrt{2 a _{max} (||{\mathbf{p}}|| - d_s)} + \frac{{\mathbf{p}}^T}{||{\mathbf{p}}||} {\mathbf{\dot p}}
    \label{CBF_point}
\end{equation}

The expression $a_{max}$ represents the highest possible acceleration value of the UAV, whereas $d_s$ is the secure distance that separates the obstacle from the UAV. Additionally, ${\mathbf{p}}$ denotes the vector from the obstacle's location to the UAV center, while ${\mathbf{\dot p}}$ represents the velocity of the UAV at a particular time instant $k$. Similarly, $h(\mathbf{x})$ can be defined as a discrete-time control barrier function (CBF). The final invariance condition can be found in equation (\ref{invariance_condition}).

\begin{equation}
\begin{aligned}
    \frac{a_{max}~ {\mathbf{\dot p}}^T {\mathbf{p}}}{\sqrt{2 a _{max} (||{\mathbf{p}}|| - d_s)}} - {\left (\frac{{\mathbf{p}}^T}{||{\mathbf{p}}||} {\mathbf{\dot p}} \right )}^{2} + ||{\mathbf{\dot p}}||^2  + {\mathbf{p}}^T \mathbf{u} \\ +  \gamma h^z(\mathbf{x}) ||{\mathbf{p}}|| ~\geq ~0
\end{aligned}
\label{invariance_condition}
\end{equation}

\subsection{Aerodynamic Ceiling, Ground and Wall effect}
When a UAV's rotors start rotating, depending on whether it is near a vertical or horizontal surface, different aerodynamic forces start acting on it. These aerodynamic forces start affecting the UAV by pulling or pushing from the expected trajectory. It is crucial to understand where these forces originate and how they affect to tackle their effects.
\newline
\subsubsection{Ground Effect}
When a UAV flies over a horizontal surface, ground effects (GE) occur. GE is an aerodynamic effect that has been studied extensively and seen to push UAVs away from the ground \cite{multirotor_forces,ge}. The theoretical model of GE presented by Cheeseman and Bennet \cite{ge_model} is a widely accepted thurst ratio approximation of GE as given in (\ref{ground_effect_eq}).

\begin{equation}
    \text{Ground Effect:} \left[ {\dfrac{T_{GE}}{T_\infty}}\right] = \dfrac{1}{1 - (\dfrac{R}{4z})^2} 
    \label{ground_effect_eq}
\end{equation}
where $T_{GE}$ is the Thrust into the Ground, $T_\infty$ is the Thrust baseline, $R$ is the radius of the propeller and $z$ is the distance from the ground.
\subsubsection{Ceiling Effect}
When a UAV flies underneath a horizontal surface, nonlinear disturbances in the form of ceiling effect (CE) acts on the UAV. Contrary to GE, CE pulls the UAV towards the surface \cite{multirotor_forces}. The mathematical approximation is found as a curve in (\ref{ceiling_effect_eq})

\begin{equation}
    \text{Ceiling Effect:} \left[ {\dfrac{T_{CE}}{T_\infty}}\right] = \dfrac{1}{1 - (\dfrac{1}{a_1})(\dfrac{R}{a_2+z})^2}
    \label{ceiling_effect_eq}
\end{equation}
where $T_{CE}$ is the Thrust into the ceiling, $a_1$ and $a_2$ are coefficients obtained through an experimental least square approach.
\subsubsection{Sidewall Effect}
 When a UAV flies close to a vertical surface, it experiences a pull toward the wall. This force is smaller than GE and CE forces and acts on the rotors randomly while pulling toward the wall, destabilizing the UAV. In \cite{force_torque}, the paper tried to model this effect and found it to be yaw invariant, but it could not model the effect as it could not detect the wall effect reliably. According to their experiments, the force along the X-Y axis varied by up to 0.052 N with a standard deviation of up to 0.022 N, and along the Z-axis varied by up to 0.062 N with a standard deviation of 0.065 N. Hence, these forces act randomly with these parameters. 
\subsubsection{Combined Tunnel effect}
The combined tunnel effect refers to the two possible combinations of aerodynamic forces acting in corners. They are near the ceiling (Ceiling effect and sidewall effect) and ground (ground effect and sidewall effect). These effects were studied in \cite{corner_effect}, as In Low Corner Effect (ILoCE) and In Upper Corner Effect (IUpCE). It  tries to analyse the effects and concluded that a source and drain vortex depicting the combined forces is formed in the corners. These vortexes are shown as Particle image velocimetry (PIV) images, and the force diagrams shows a higher combined force in the corners than the individual forces.

\section{PROBLEM FORMULATION}

\begin{figure}
    \centering
    \includegraphics[width=0.48\textwidth]{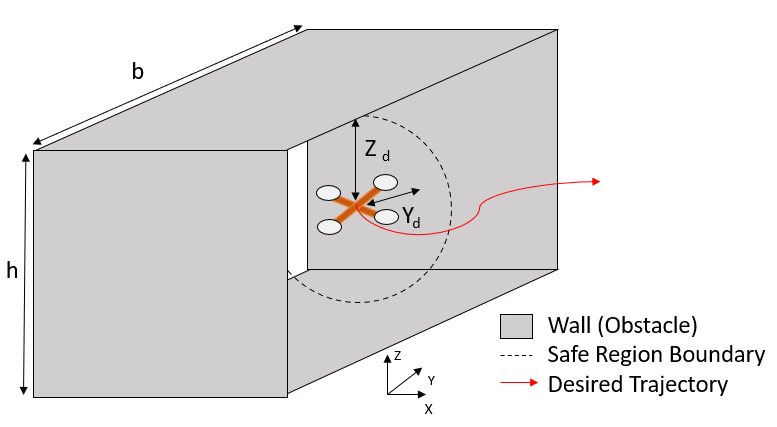}
    \caption{Image shows the trajectory tracking of a UAV inside a tunnel while handling effects from tunnel effects.}
    \label{problem_def_2}
\end{figure}

The aim of the paper is to provide a control strategy to avoid tunnel effects (combined, ceiling, sidewall and ground effects) and tackle their disturbance to provide safety guarantees when the UAV is close proximity to the tunnel walls as shown in Fig. \ref{fig1} and Fig. \ref{problem_def_2}. These tasks have been defined in 3 cases. We consider UAV center to be same as the UAV center of gravity.

\subsubsection{\textbf{Case I}}
To follow trajectory inside a tunnel while maintaining a minimum distance of $2~ \times$ Radius of Propeller (Fig. \ref{fig1}) to avoid aerodynamic interactions with the wall. The UAV will be bound inside a safe region of operation (Negligible Aerodynamic interactions) in the presence of external disturbances in the form of wind. 

\subsubsection{\textbf{Case II}}
Minimize the safe distance of operation ($z_d$, $y_d$, $h - z_d$ and $b - y_d$) (Fig. \ref{problem_def_2}) from the tunnel walls for close proximity operations. We minimize the safe hovering distance from the walls even in the presence of external disturbances. 

\subsubsection{\textbf{Case III}}
To follow a trajectory with close proximity to the wall, ceiling and ground and tackle the combined tunnel aerodynamic effect.

Primary objective in trajectory tracking and hovering tasks is defined as the error ($e(\mathbf{x}))$ in (\ref{trajectory_error}).

\begin{equation}
     \underset{\mathbf{u}} {\min} \ e(\mathbf{x}) =  ||\mathbf{p}(\mathbf{x})- \mathbf{p}^{d}|| \  \forall \ k > 0
     \label{trajectory_error}
\end{equation}

where $p(\mathbf{x})$ is the position of UAV center, $p^d$ is the desired position of the UAV center and $k$ is the discreet time step.

\section{PROPOSED CONTROLLER}

\begin{figure*}[ht]
\includegraphics[scale=0.48]{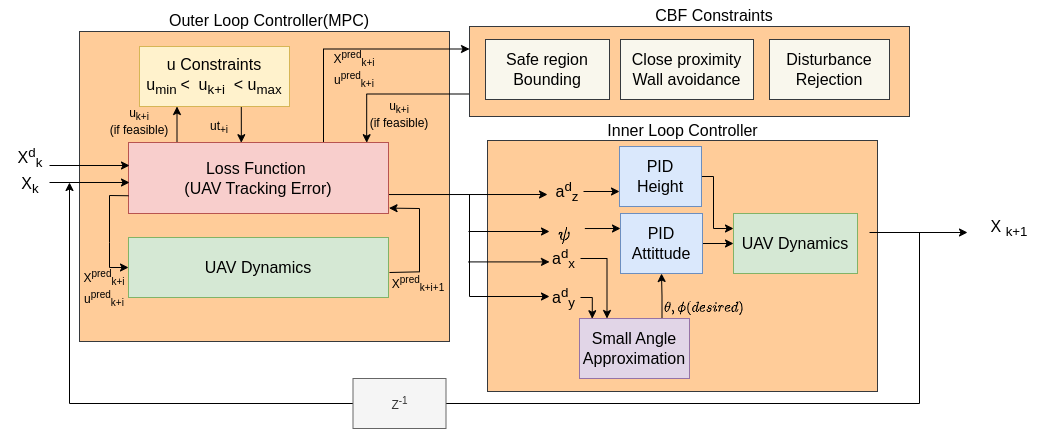}
\caption{\small{\textbf{Control Architecture:} The Outer loop control for the UAV is a Model Predictive controller which provides the optimal control input to the Inner loop control (PID) to track while additional constraints to the MPC are derived from the CBF. $x^{d}_k$ is the desired state of the UAV. }}
\label{mpc_diagram}
\end{figure*}

The control architecture of the proposed controller is presented in Fig. \ref{mpc_diagram}. The control loop consists of an outer loop Model Predictive Control (MPC) with safety constraints derived from a modified Control Barrier Function (CBF). The modifications to CBF are made to restrict the UAV inside a desired safe region contrary to its earlier collision avoidance utility. A disturbance rejection term is also introduced to the conventional CBF to handle the Tunnel effect and other wind disturbances in the tunnel. 
The inner loop control is comprised of thrust and attitude PID control. We present our main contributions in this section. We begin by writing the discrete-time dynamics of the UAV for calculating the cost inside MPC outer loop. The state vector of the UAV is defined as $\mathbf{x_k} = [\mathbf{p_k}, \mathbf{\dot p_k}, \Psi_k, \dot \Psi_k]$ where $\mathbf{p_k}$ is the position of the UAV center in the inertial frame and $\Psi_k$ is the yaw angle of the UAV. The control input is $u_k = [\mathbf{\ddot p_k}, \ddot \Psi_k]$ at time interval $k$. The state space model for the UAV utilized by the model MPC is given in (\ref{eq_control_affine}).

\subsection{Model Predictive Control (Outer loop)} 

The optimal control problem for each time step $k$ is given in (\ref{optimal_problem}) for the UAV dynamics.

\begin{subequations}
\label{eq:Cost_function}
\begin{align}
u^{opt}_k = \underset{\mathbf{u}} {\min} \ g( \mathbf{x_k,u_k}, t_{k})\\
\textrm{s.t.} \  \mathbf{\dot x}_{k} = \mathbf{A} \mathbf{x}_{k} + \mathbf{B} \mathbf{u}_{k}\\
\mathbf{x}_{min} \leq \mathbf{x}_k \leq \mathbf{x}_{max}\\
\mathbf{u}_{min} \leq \mathbf{u}_k \leq \mathbf{u}_{max}
\end{align}
\label{optimal_problem}
\end{subequations}

$u^{opt}_k$ depicts the optimal input by the optimizer which is then given to the inner loop controller for tracking. The cost $g$ is the weighted sum of $N_g$ cost functions $g = \sum_{i = 1}^{N_g} g_i$ given below. $N_g$ is the number of cost functions and $N$ denotes the prediction horizon of the MPC.

\subsubsection{UAV center tracking error}

To account for the penalization of drift from the desired position or trajectory, we add a cost to the MPC optimizer as in (\ref{trajectory_tracking_cost})

\begin{equation}
\begin{aligned}
g_{1} = \ \sum_{i = 0}^{N-1} (  \left|\left|  \mathbf{p}(\mathbf{x_{k+i}})- \mathbf{p}^{d}_{k+i}  \right|\right|^2_{W_{1}} ) + \left|\left|  \mathbf{p}(\mathbf{x_{k+N}})- \mathbf{p}^{d}_{k+N}  \right|\right|^2_{W_{s_{1}}}
\end{aligned}
\label{trajectory_tracking_cost}
 \end{equation}

where $W_1$ and $W_{s_1}$ are weight matrices. 

\subsubsection{UAV center velocity error}

To penalize the higher velocity of the UAV, we add a cost to the MPC optimizer as in (\ref{velocity_cost}).

\begin{equation}
\begin{aligned}
g_{2} = \ \sum_{i = 0}^{N-1} (  \left|\left|   \mathbf{\dot p}(\mathbf{x}_{k+i})   \right|\right|^2_{W_{2}} ) + \left|\left|  \mathbf{\dot p}( \mathbf{x}_{k+N})  \right|\right|^2_{W_{s_{2}}}
\end{aligned}
\label{velocity_cost}
\end{equation} 

where $W_2$ and $W_{s_2}$ are weight matrices.

The controller with only MPC as the outer loop and PID as the inner loop is referred to as \textbf{Naive MPC} in the following sections. \textbf{MPC-HC} is demonstrated as Naive MPC with hard constraints on the optimizer, not in the form of CBF. These algorithms would be utilized to compare the performance of the proposed controller. Additional constraints for \textbf{MPC-HC} are: For \textbf{Case I}, $||\mathbf{p} - d^s|| \leq r$ and for \textbf{Case III}, $||\mathbf{d}|| \geq d_s$ which have been explained in Section VI part C.

\subsection{PID (Inner loop)}

The inner loop PID receives a $u^{opt}_k$ as the optimal $u_k$ from the MPC optimizer. Desired roll $\Theta$ and pitch $\Phi$ angles are calculated using small angle analysis, and the desired thrust and attitude are tracked by PID Thrust and Attitude Controllers. 

\subsection{CBF Constraints}

\subsubsection{Bounding UAV in safe region (Bounding condition)}

For \textbf{Case I}, We give our primary contribution to bound the UAV inside the safe region where aerodynamic effects do not hamper the stability of the UAV. To assume a continuous differentiable bounding area, we choose the safe region to be a spherical boundary similar to Fig. (\ref{fig1}), as the tunnel effect and other effects together form a region where the safe region can be simplified to a sphere. The UAV can only leave the safe region in a radial direction. We constrain the movement of the UAV for a high velocity motion using CBF. The CBF for one direction is given in \ref{h_bounding}, and we replace $\mathbf{p}$ with $-\mathbf{p}$ to get CBF in the opposite direction.

\begin{equation}
h_{1}(\mathbf{x}_k) = \sqrt{\frac{2 ({\mathbf{p}}_{k+i})^T a_{max}}{||{\mathbf{p}}_{k+i}||}  (||{\mathbf{p}}_{k+i}|| - r)} + \frac{{\mathbf{p}}_{k+i}^T}{||{\mathbf{p}}_{k+i}||} ({\mathbf{\dot p}}_{k+i} - {\mathbf{\dot p}}^t_{k+i})
\label{h_bounding}
\end{equation}

Where $r$ is the radius of the safe region.

\subsubsection{Minimize safe distance of operation from tunnel Walls (Disturbance Rejection)}

For \textbf{Case II}, we can tighten the bound of the CBF using an additional disturbance rejection parameter $\lambda$ to tackle aerodynamic disturbances from various effects. Hence we change the earlier invariance condition in (\ref{invariance}) to the condition in (\ref{disturbance_invariance}).

\begin{equation}
    \dot{h}(\mathbf{x}) + \gamma({h}^z(\mathbf{x}) - \lambda) \geq 0  
    \label{disturbance_invariance}
\end{equation}

\subsubsection{Trajectory tracking for close proximity flights}

For \textbf{Case III}, the CBF is modified to avoid walls and the CBF condition for this task is given in (\ref{cbf_wall}).

\begin{equation}
\noindent h_2(\mathbf{x}_k) = \sqrt{\frac{2 {\mathbf{d}}(\mathbf{x}_{k+i})^T a_{max}}{||{\mathbf{d}}(\mathbf{x}_{k+i})||}  (||{\mathbf{d}}(\mathbf{x}_{k+i})|| - d^{s})} + \frac{{\mathbf{d}}(\mathbf{x}_{k+i})^T}{||{\mathbf{d}}(\mathbf{x}_{k+i})||} {\mathbf{\dot p}}_{k+i} 
\label{cbf_wall}
\end{equation}

where $d(\mathbf{x}_{k})$ is the perpendicular distance from the wall at time instance $k$ and $d^{s}$ is the minimum safe distance from the wall. The combination

\begin{figure}[hbt!]
    \subfigure[] {\includegraphics[width=0.49\columnwidth]{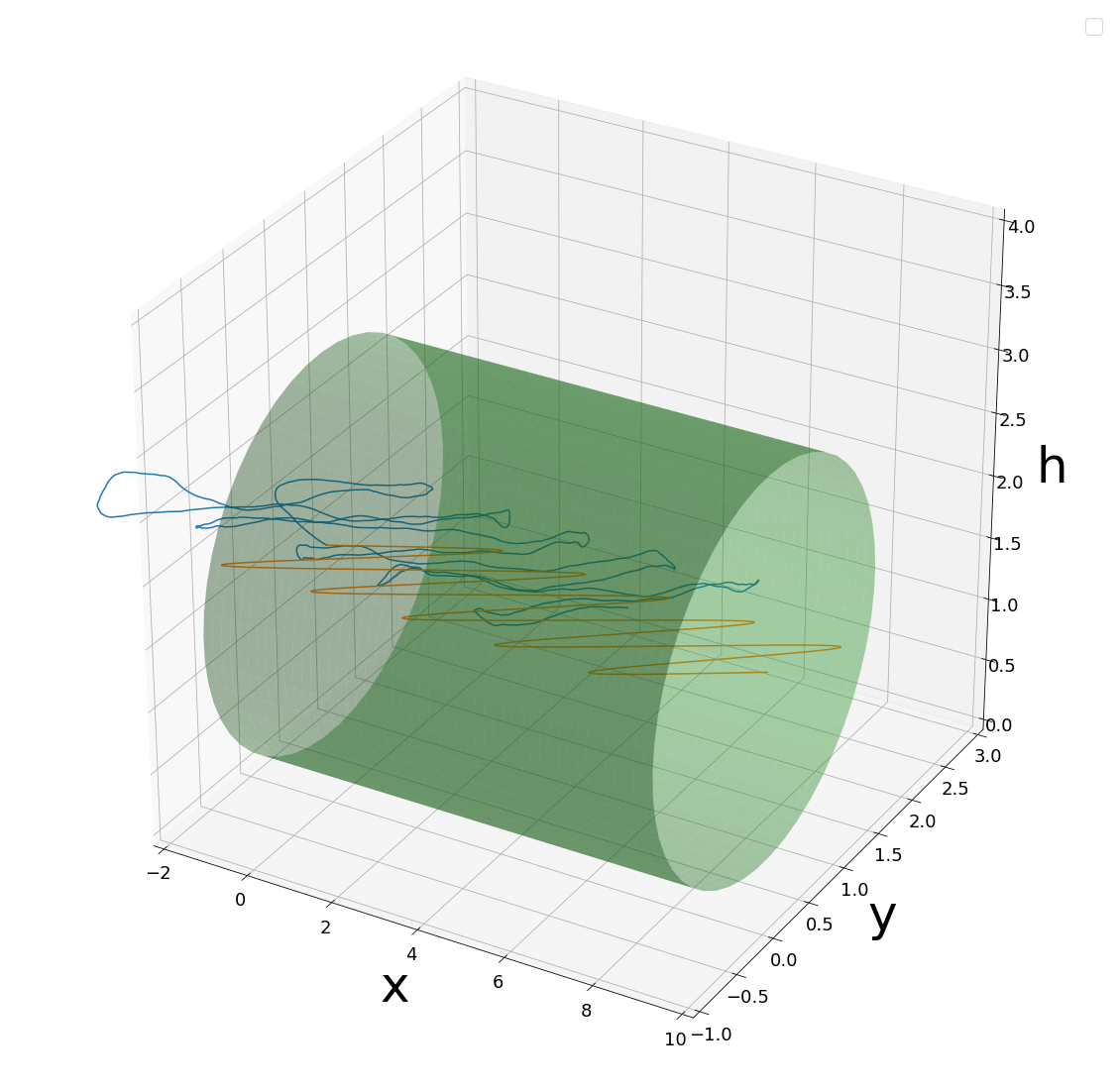}}
    \subfigure[] {\includegraphics[width=0.49\columnwidth]{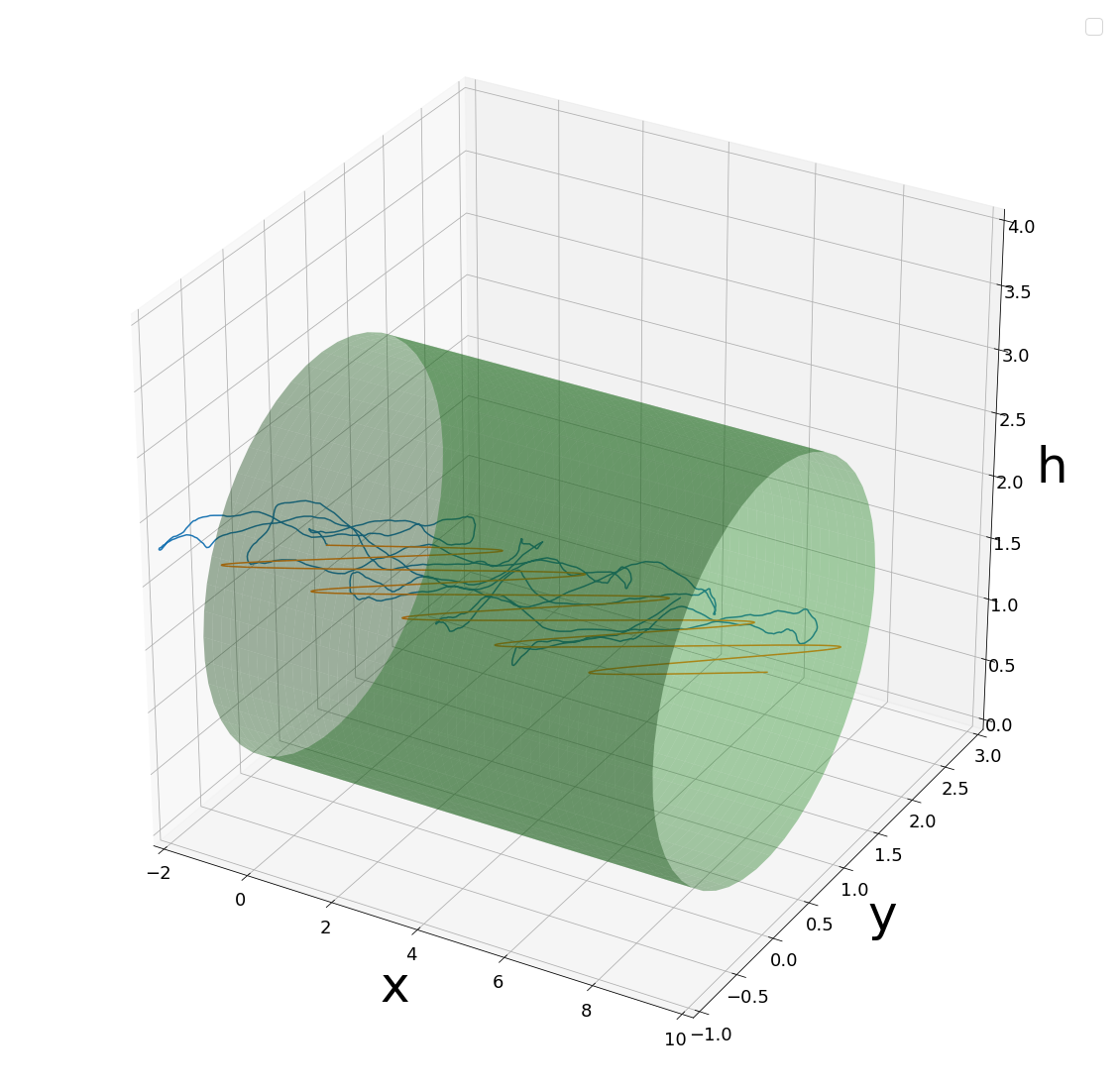}}
    \subfigure[] {\includegraphics[width=0.49\columnwidth]{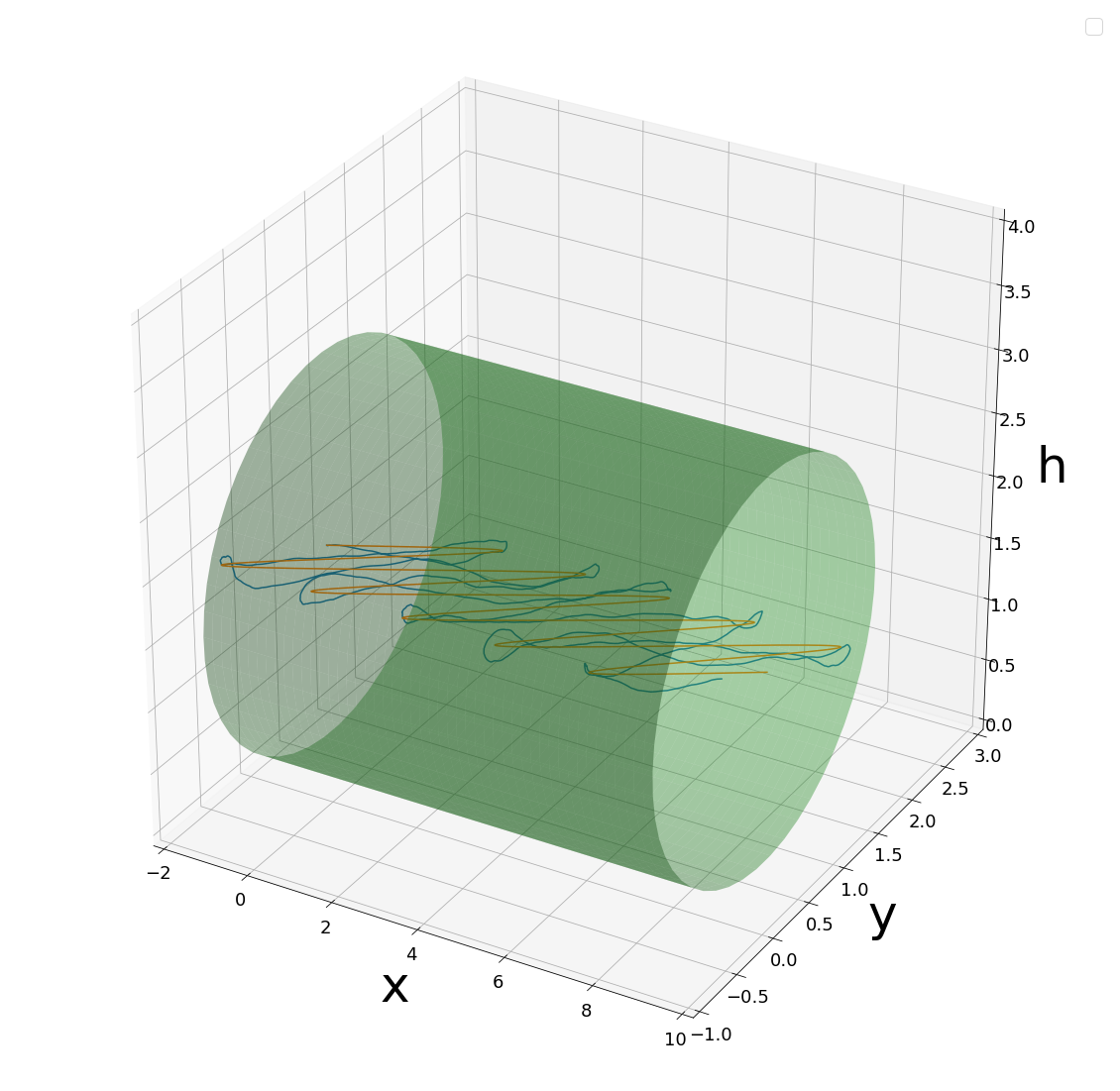}}
    \caption{{Position of the UAV center and the safe region boundary (\textbf{Case I}) for trajectory tracking in Fig. \ref{fig1}: } (a) Naive MPC, (b) MPC-HC, (c) MPC- CBF. <green> -> safe region boundary, <blue> -> UAV center, <orange> -> desired trajectory. UAV Center should remain bound inside the safe region.}
    \label{MPC_Bounding}
\end{figure}

\section{Simulation Results}

This section presents the results of the performance of the algorithm on simulation.
Python 3 was used to perform the scenario on an Intel® Core™ i7-8550U CPU desktop operating at 1.80 GHz. The optimizer used is the 'SLSQP' method provided in the scipy library \cite{Virtanen_2020}. The specifications of the UAV model used are given in Table \ref{UAV_table} and the parameters used in \textbf{MPC-CBF} are given in Table. \ref{MPC_table}.
 
\begin{table}[!htbp]

\centering
\resizebox{\columnwidth}{!}{%
\begin{tabular}{|>{\centering}p{0.17\textwidth}|>{\centering\arraybackslash}p{0.26\textwidth}|}
\hline
\textbf{Parameter} &  \textbf{Value} \\
\hline
Mass &   $1.5~kg$ \\
\hline
Arm length & $0.20~m$\\
\hline
Propeller Diameter & $0.24~m$ \\
\hline
Moment of Inertia - UAV&  $I_x = 0.1~kg~m^2$, $I_y = 0.1~kg~m^2$, $I_z = 0.2~kg~m^2$\\
\hline
UAV attitude constraints & $|\theta| \leq \pi/10~ rad$ , $|\phi| \leq \pi/10~ rad$\\ 
\hline
\end{tabular}}
\caption{Specifications of the UAV: These parameters have been taken from the UAV used to define the Aerodynamic effects}
\label{UAV_table}

\end{table}

\begin{table}[!htbp]
\begin{center}
{\begin{tabular}{|>{\centering}p{0.15\textwidth}|>{\centering\arraybackslash}p{0.28\textwidth}|}
\hline
\textbf{Parameter} &  \textbf{Value} \\
\hline
MPC Weights &   $w_1 = 10\times \mathbf{I}_{3\times3}$, $w_{s_1} = 50\times \mathbf{I}_{3\times3}$, $w_2 = 2\times \mathbf{I}_{3\times3}$, $w_{s_2} = 10\times \mathbf{I}_{3\times3}$ \\
\hline
$u_k$ Initialization & $\mathbf{0}_{1 \times 4n}$ \\
\hline
$\gamma$ & $3$  \\
\hline
$\lambda$ & $8$  \\
\hline
$z$ & $3$  \\
\hline
Sampling step ($t_s$) & $0.1$ s \\
\hline
Total time ($t$) & $100$ s \\
\hline
Max wind disturbance & $d_m  = 0.8\, m/s^2$ \\
\hline

\hline
\end{tabular}}
\caption{Weights and Parameters for MPC and CBF}
\label{MPC_table}
\end{center}
\end{table}

\subsection{Metric for performance comparison}

We measure the performance of the algorithm with the following matrices:

\begin{itemize}
    \item Bounding inside Safe region 
    \item Trajectory Tracking error, $T_e = \sqrt{\frac{1}{N} \sum_{k = 0}^{N-1} (\mathbf{p}(\mathbf{x}_{k}) - p^d_k)^2}$
    \item Control effort, $c_e = \sum_{k = 0}^{N-1}{||\mathbf{u}_k||^2}$ 
    \item Control Smoothness, $c_s = \sum_{k = 0}^{N-1}{|\Delta\mathbf{u}_k|}$ 
\end{itemize}

\subsection{Results for \textbf{Case I}}

For bounding the UAV inside the safe region, \textbf{Naive MPC} is unable to find the bounds and shows very high Trajectory tracking error in the presence of wind disturbances. \textbf{MPC - HC} is unable to maintain the bound when the UAV gets a high velocity input. \textbf{MPC-CBF} performs best compared to other algorithms because it incorporates obstacle avoidance and disturbance rejection using CBF. It shows a 30\% decrease in the trajectory error and maintains the safe region's bound. Trajectory tracking results are shown in Fig. \ref{MPC_Bounding_trajectory} with Fig. \ref{MPC_Bounding} depicting the trajectory in 3D. The performance matrices are mentioned in Table. \ref{performance_matrix}.

\subsection{Results for \textbf{Case II}}

The shortest distance between the walls and the UAV depicts the extended stability zone of the UAV when deploying a new control algorithm. \textbf{Naive MPC} gives the minimal distance as $2 \times$ R while \textbf{MPC-CBF} shows a decrease in this distance by 45\% as shown in Table. \ref{performance_matrix}.

\subsection{Results for \textbf{Case III}}

When the UAV trajectory passes through the unsafe region, \textbf{Naive MPC} and \textbf{MPC-HC} are unable to maintain the trajectory and subsequently collide to the wall. Only \textbf{MPC-CBF} is able to maintain the trajectory while reducing the control effort by \~15\% thus reducing the power consumed by the UAV.

\begin{figure}[!htbp]
    \subfigure[] {\includegraphics[width=0.93\columnwidth]{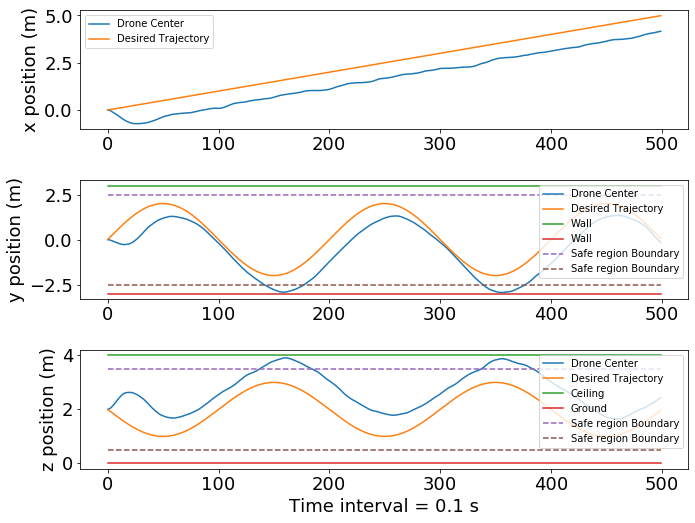}}
    \subfigure[] {\includegraphics[width=0.93\columnwidth]{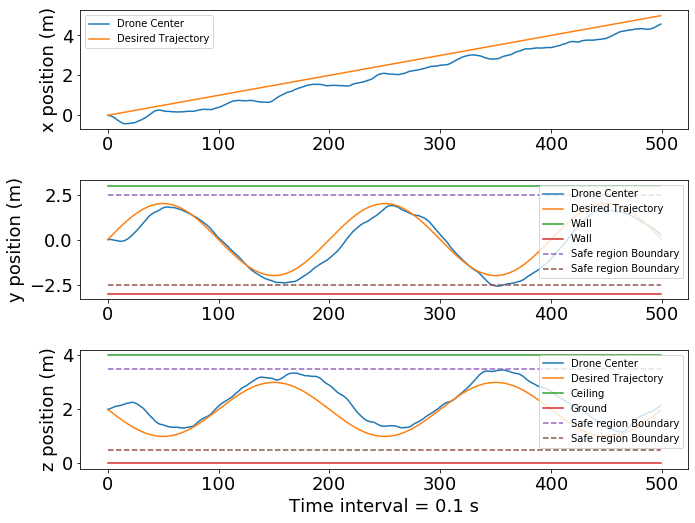}}
    \subfigure[] {\includegraphics[width=0.93\columnwidth]{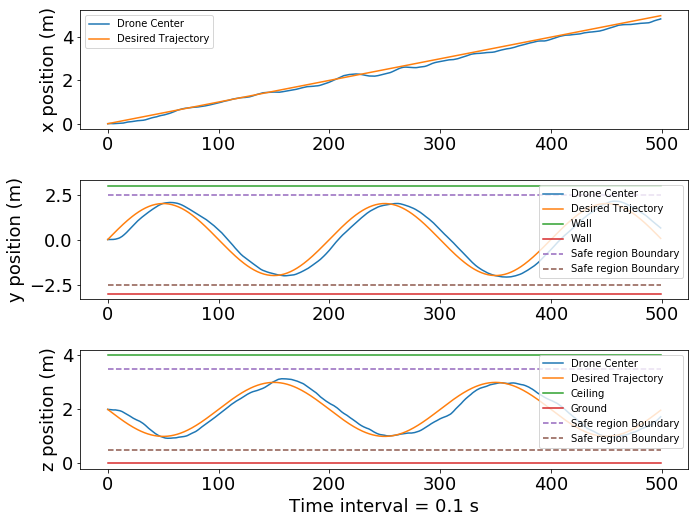}}
    \caption{\textbf{UAV position to maintain safe region (Case I)} (a) Naive MPC, (b) MPC-HC, (c) MPC- BLF. It shows that the UAV leaves the safe region for Naive MPC and MPC-HC but maintains the safe region for MPC-CBF. }
    \label{MPC_Bounding_trajectory}
\end{figure}

\begin{figure}[!htbp]
    \subfigure[] {\includegraphics[width=0.93\columnwidth]{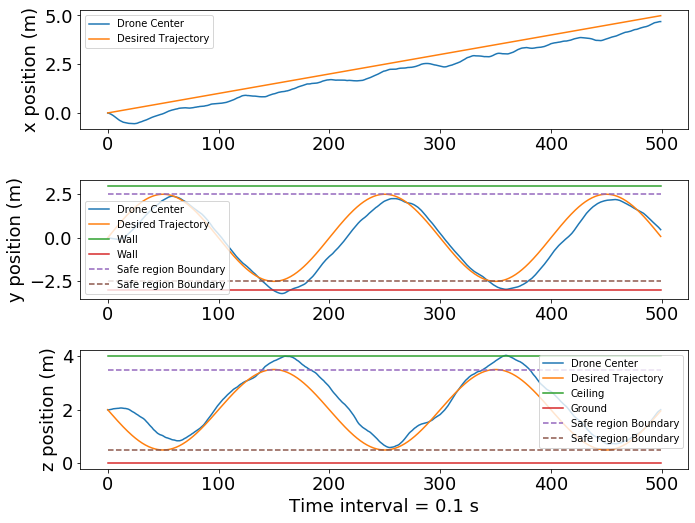}}
    \subfigure[] {\includegraphics[width=0.93\columnwidth]{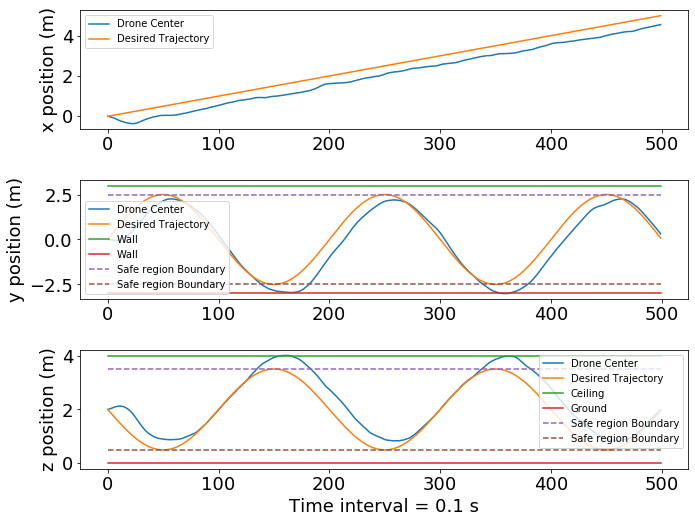}}
    \subfigure[] {\includegraphics[width=0.93\columnwidth]{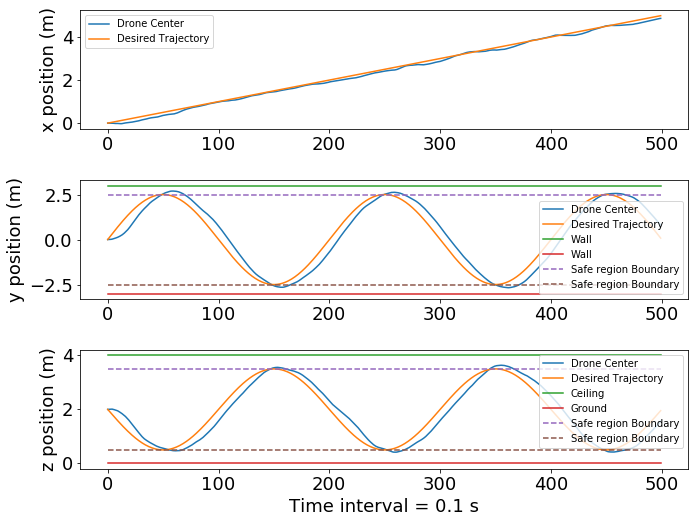}}
    \caption{\textbf{UAV position while trajectory in unsafe region (Case III)} (a) Naive MPC, (b) MPC-HC, (c) MPC- BLF. It shows that the UAV collides with the wall for Naive MPC and MPC-HC, but maintains the trajectory for MPC-CBF. }
    \label{MPC_Bounding_trajectory}
\end{figure}

\begin{table*}[!htbp]
\centering
\begin{tabular}{ |p{4cm}|p{3cm}|p{2.7cm}|p{2.7cm}|p{2.7cm}|  }
\hline
\multicolumn{2}{|c|}{UAV Tasks} & Naive MPC& MPC - HC  & MPC - CBF \\
\hline
\multirow{4}{*}{\parbox{3cm}{Bounding in safe region (\textbf{Case I})}} & {Maintain Boundary} & {$\times$} & {$\times$}& {$\textbf{\checkmark}$}\\ 
\cline{2-5}& {$T_e (m)$} & {1.16250} & {0.70152}& {\textbf{0.51087}}\\ 
\cline{2-5}& {$c_e$} & {0.95237} & {0.93959}& {\textbf{0.74246}}\\  
\cline{2-5}& {$c_s$} & {0.09519} & {0.12497}& {\textbf{0.09190}} \\

\hline

\multirow{3}{*}{\parbox{3cm}{Minimum distance to wall (\textbf{Case II})}} & {Ground effect (m)} & {0.495} & {0.521}& {\textbf{0.312}}\\ 
\cline{2-5}& {Ceiling effect (m)} & {0.502} & {0.478} & {\textbf{0.298}}\\ 
\cline{2-5}& {Sidewall effect (m)} & {0.481} & {0.465} & {\textbf{0.138}}\\ 
\hline
\multirow{4}{*}{\parbox{3cm}{Close proximity trajectory tracking (\textbf{Case III})}} & {Collision} & {$\checkmark$} & {$\checkmark$}& {$\mathbf{\times}$}\\ 
\cline{2-5}& {$T_e (m)$} & {1.38771} & {0.8327} & {\textbf{0.56010}}\\ 
\cline{2-5}& {$c_e$} & {1.11998} & {0.99483}& {\textbf{0.89015}}\\  
\cline{2-5}& {$c_s$} & {0.10604} & {0.09343} & {\textbf{0.06222}}\\ 
\hline
\end{tabular}
\caption {\textbf{Algorithm benchmarking:} We compare the Trajectory rms error, control effort and control smoothness of MPC-CBF while flying amidst external disturbances with other algorithms, and it performs substantially better than all other algorithms.}
\label{performance_matrix}
\end{table*}




\vspace{2cm}

\section{CONCLUSION}

The paper shows that a Model predictive controller, when combined with constraints using Control Barrier Function, can provide safety guarantees when flying inside a tunnel. The controller also reduces the safe hovering distance from the wall by 37\% and incorporates high disturbance tolerance.  
It is also shown that flying near the ground and ceiling can reduce the UAV's power consumed (control effort) by \~ 15\%. The algorithm's efficacy provides safety guarantees while travelling inside a tunnel using parameters from a real UAV model.
Future work shall include using vision based learning models to detect obstacles and create barrier functions through their understanding.


\bibliographystyle{IEEEtran}
\bibliography{IEEEabrv, references}

\end{document}